\documentclass[grid,balance,upint,subscriptcorrection,varvw,mathalfa=cal=euler,spanish,french,greek,russian,vietnamese, nofoot]{asmeconf}

\graphicspath{{images/}}

\hypersetup{%
	pdfauthor={Siemen Herremans},                       		      
	pdftitle={ASME Conference Paper LaTeX Template},                  
	pdfkeywords={ASME conference paper, LaTeX template, BibTeX style},
	pdfsubject = {Describes the asmeconf LaTeX template with author grid},
	pdflicenseurl={https://ctan.org/pkg/asmeconf},
}

\usepackage{bbm}
\usepackage{longtable}
\usepackage{fancyhdr}
\usepackage{supertabular}

\begin{document}

\DeclareRobustCommand{\rchi}{{\mathpalette\irchi\relax}}
\newcommand{\irchi}[2]{\raisebox{\depth}{$#1\chi$}}




\title{Autonomous Port Navigation with Ranging Sensors using Model-Based Reinforcement Learning} 
 
%
%
%


\SetAuthors{%
	Siemen Herremans \affil{1}\CorrespondingAuthor{siemen.herremans@uantwerpen.be}, 
	Ali Anwar\affil{1}, 
	Arne Troch \affil{1},
	Ian Ravijts \affil{1},
	Maarten Vangeneugden \affil{3},
	Siegfried Mercelis \affil{1},
	Peter Hellinckx \affil{2}
}

\SetAffiliation{1}{IDLab, University of Antwerp - imec\\ Antwerp, Belgium}
\SetAffiliation{2}{IDLab, University of Antwerp\\ Antwerp, Belgium}
\SetAffiliation{3}{University of Ghent\\ Ghent, Belgium}

\lhead{Published in the conference proceedings of ASME OMAE 2023. June 11 - 16, 2023. Melbourne, Australia. Author version.}


\maketitle

\versionfootnote{Preprint}




\begin{abstract}
Autonomous shipping has recently gained much interest in the research community. However, little research focuses on inland - and port navigation, even though this is identified by countries such as Belgium and the Netherlands as an essential step towards a sustainable future. These environments pose unique challenges, since they can contain dynamic obstacles that do not broadcast their location, such as small vessels, kayaks or buoys. Therefore, this research proposes a navigational algorithm which can navigate an inland vessel in a wide variety of complex port scenarios using ranging sensors to observe the environment. The proposed methodology is based on a machine learning approach that has recently set benchmark results in various domains: model-based reinforcement learning. By randomizing the port environments during training, the trained model can navigate in scenarios that it never encountered during training. Furthermore, results show that our approach outperforms the commonly used dynamic window approach and a benchmark model-free reinforcement learning algorithm. This work is therefore a significant step towards vessels that can navigate autonomously in complex port scenarios.
\end{abstract}

\newpage
\begin{nomenclature}
\EntryHeading{Set Operations}
\entry{$\mathcal{K} \times \mathcal{P}$}{The Cartesian product of two sets $\mathcal{K}$ and $\mathcal{P}$.}
\entry{$\mathcal{K}^n$}{The Cartesian power, defined as $\underbrace{\mathcal{K} \times \mathcal{K} \times ... \times \mathcal{K}}_{n}$.}

\EntryHeading{Vector Operations}
\entry{$\mathbf{a}\mathbin\Vert\mathbf{b}$}{The concatenation of two vectors $\mathbf{a}$ and $\mathbf{b}$.}
\entry{$dim(\mathbf{a})$}{The amount of dimensions of vector $\mathbf{a}$.}
\entry{$\left\lVert \mathbf{a} \right\rVert_{1}$}{The $L_1$-norm of vector $\mathbf{a}.$}

\EntryHeading{Special functions}
\entry{$\mathbbm{1}_{\mathcal{K}}(k)$}{Indicator function: 1 if $k \in \mathcal{K}$, else 0.}




\EntryHeading{Probability}
\entry{$\mathbb{E}_d$}{Expectation under distribution $d$.}
\entry{$\mathpalette{U}(x_{min}, x_{max}$)}{A uniform distribution between $x_{min}$ and $x_{max}$.}




\end{nomenclature}


\section{Introduction}


A recent study from the European commission \cite{eucomm} finds that inland waterway transport (IWT) has a much lower carbon footprint compared to other modes of transport with only half the amount of $CO_2/(ton*km)$ compared to railroad transport and only 17$\%$ that of road based transport. Additionally, IWT has the potential to decongest road networks, which improves supply chain reliability and further lowers the carbon footprint of the European Union. The ports of Rotterdam and Antwerp are the two largest ports in Europe, combining this with the fact that both The Netherlands and Belgium have densely connected inland waterways, these countries can benefit greatly from IWT to transport freight from the port throughout the country. However, due to a shortage of skippers in Belgium and The Netherlands, it is not possible to increase the reliance on IWT in the supply chain without a form of \mbox{(semi-)autonomous} vessel control. Also, a large part of vessel collisions are caused by human error, of which a significant portion has the potential to be avoided with autonomous shipping \cite{safetyshipping}. 

Therefore, research interest in autonomous shipping has increased \cite{shippingreview}, major advancements can partly be attributed to the application of recent advancements in deep learning methods such as generative models and convolutional networks \cite{shippingvision} or deep reinforcement learning (RL) \cite{shipping_rl1} \cite{shipping_rl2}. RL is a methodology where an agent must learn a behavior by maximizing a reward signal in an environment \cite{sutton_barto}; this method allows for superhuman performance, since it does not rely on a human-labeled data set such as supervised deep learning methods. For a long time, the application potential of RL was limited to a small problem set. However, recent advancements in the field have generalized this methodology to a wider set of problems with complex observational inputs. These advancements are often demonstrated by learning to play video games without human data. 


Reinforcement learning approaches can roughly be split into two major categories: model-based and model-free. Model-free reinforcement learning uses its experience to either learn an evaluation function of how desirable certain actions are in a certain state (Q-learning) \cite{dqn} or directly learns a distribution that predicts the likelihood of an action being optimal in a state (policy optimization) \cite{ppo}. Notably, these methods never learn the actual dynamics of the environment. Thus, they do not have the ability to predict the outcome if they perform a certain action in the environment, they only learn a heuristic of how good the action is expected to be. On the contrary, model-based RL (MBRL) learns that predictive capability, which allows these algorithms to combine model-free RL with some form of planning over the (learned) dynamics model \cite{mbrl_survey}. This capability has been linked to various theoretical advantages such as explainability, stability and most notably: data-efficiency \cite{mbrl_survey}. 
MBRL was for a long time outperformed by its model-free counterparts. However, recent works have also extended this methodology to a wider range of problems; they show that MBRL can match or outperform model-free methods on a variety of tasks while being significantly more data-efficient \cite{muzero} \cite{mb_mpo}. This paper builds upon these advancements and propose a methodology that extends previous work in autonomous shipping (see Sect. \ref{sec:related}) with the use of a MBRL agent. 

The main goal of this paper is threefold. First, we propose a methodology of approaching optimal path planning as a reinforcement learning problem. An algorithm is demonstrated that can combine path following with collision avoidance of dynamic obstacles. Here, ranging sensors are used to observe the environment. Second, we show the ability of reinforcement learning to generalize over a large space of scenarios by training the algorithm in various randomized scenarios. The performance of the path planning is validated on scenarios that were never encountered during training. Finally, we demonstrate that our MBRL approach outperforms both the dynamimic window approach \cite{dwa_shipping} and a state-of-the-art model-free reinforcement learning algorithm.


This work first (Sect. \ref{sec:related}) relates this work with the current state of the art, both in autonomous shipping and RL; the added value of this work is also highlighted in this section. Then, Sect. \ref{sec:formulation} proposes a theoretical formulation of optimal path planning as a reinforcement learning problem, where it is shown that an optimal RL agent is able to provide optimal paths in environment with dynamic obstacles. Sect. \ref{sec:methodology} thoroughly describes our approach and justifies our design choices. Finally, the results of our methodology are examined with qualitative and quantitative measurements in Sect. \ref{sec:results}.

\section{Related works} \label{sec:related}
Firstly, some works in the autonomous shipping literature focus purely on path following, assuming that the global path never hits an obstacle. The main focus of these works is accurate path tracking with complex dynamical simulations and external disturbances. A recent work by Wang et al. \cite{fm_rl} combines the Fast Marching \cite{fast_marching} method with RL to provide path following behavior under disturbances from wind and waves. Fossen et al. \cite{los} proposed a methodology to achieve path following on underactuated vessels. Good performance was demonstrated in simulation and on a 1:70 scale vessel. Peng et al. \cite{los_planning} built upon that methodology to achieve path following by using MPC to optimize over a learned model.

Secondly, many works focus on COLREG-compliant collision avoidance in maritime scenarios. These works assume that the only obstacles are other vessels and that their positions are perfectly known. Some notable examples are described next. Abdelaal et al. \cite{nmpc} proposed a non-linear MPC methodology to control where paths that end up in a collision within the prediction horizon are filtered out. However, perfect information about the position of obstacles and the dynamics model is assumed. Kim et al. \cite{dwa_shipping} employ the Dynamic Window Approach (DWA) to perform collision avoidance at sea. Chun et al. \cite{shipping_rl1} employ an RL algorithm for collision avoidance and COLREG compliance at sea. They limit the scope to avoiding other vessels that share their current position and velocity. Their proposed methodology successfully avoids safety violations in scenarios with many other vessels. Jiang et al. \cite{colav_attention} build upon this work by adding an attention mechanism to the RL system and also provide COLREG compliant collision avoidance in maritime environments. This attention mechanism is a machine learning technique which improves the ability of the algorithm to decide on which other vessels it should focus, depending on the current context. Their work makes similar assumptions as Chun et al., but it also remains limited to rudder control only. 

Thirdly, some works take steps towards autonomous shipping in scenarios with imperfect information, such as vessels that do not share their position or (drifting) buoys. Vanneste et al. \cite{smart_waterways} proposed model-predictive reinforcement learning and showed that their approach provided safer paths compared to model-predictive control (MPC) in the Frenet frame. Similar to our work, they also employed ranging sensor observations to deal with unknown obstacles. Zhang et al. \cite{apf_rl} provide a system that relates the closest to this work. They provide an algorithm that is able to navigate in scenarios with unknown static and dynamic obstacles, however, their algorithm only works in one specific port/waterway and is therefore not a truly general approach. Finally, Chen et al. \cite{cnn_smallships} address the necessity of observation-based autonomous navigation but limit the work to improving the observations without proposing a navigation methodology.

All of the previously mentioned contributions are summarized in table \ref{tab:related}.

\begin{center}

\topcaption{A comparison of our work with different autonomous shipping \mbox{approaches}.}

\tablefirsthead{
    \hline \multicolumn{1}{|c|}{\textbf{Contribution}} & \multicolumn{1}{|c|}{\textbf{\shortstack{Path \\ Following}}} & \multicolumn{1}{|c|}{\textbf{\shortstack{Avoidance\\Type}}} & \multicolumn{1}{|c|}{\textbf{{\shortstack{Imperfect \\ Information}}}} \\ \hline 
}

\begin{supertabular}{|l|l|l|l|} \label{tab:related} 


Wang et al. \cite{fm_rl} & Yes & None & No \\
Fossen et al. \cite{los} & Yes & None & No \\
Peng et al. \cite{los_planning} & Yes & None & No \\
\hline
Chun et al. \cite{shipping_rl1} & Yes & Dynamic & No \\
Jiang et al. \cite{colav_attention} & Yes & Dynamic & No \\ 
Abdelaal et al. \cite{nmpc} & Yes & Dynamic & No \\
Kim et al. \cite{dwa_shipping} & Yes & Dynamic & No \\
\hline
Vanneste et al. \cite{smart_waterways} & Yes & Static & Yes  \\
Zhang et al. \cite{apf_rl} & No & Stat.+Dyn. & Yes \\
\hline
\emph{Our contribution} & Yes & Stat.+Dyn. & Yes  \\
\hline

\end{supertabular}

\end{center}

\section{Problem Formulation} \label{sec:formulation}

This section first describes the problem of optimal path planning in a formal manner, where we follow the formalism described in \cite{rrt_star}. Global (offline) planning is differentiated from (online) local planning. We then show that it is possible to formulate the local planning problem as a Markov Decision Process (MDP), optimizing a behavior in this MDP provides an optimal local path. Finally, it is noted that RL is a method to approximate the solution of an MDP.

\subsection{Autonomous Navigation}
Autonomous navigation can be split into four tasks: global planning, local planning, path following and control. Global planning provides a path from the current position to the destination without considering dynamic obstacles; this can be precomputed using only a map of the waterway. Local planning employs real-time observations with the purpose of following the global path while taking additional constraints into account, such as avoiding dynamic obstacles. This means that an autonomous navigation methodology first computes a global path to find a route from the start to the goal. That global path is then used by the local planner to safely guide the vessel to the destination. Subsequently, the local path is then followed as closely as possible by defining the correct setpoints for the actuators of the vessel. In this work, the actuators are the rudder (angle) and the propellor (thrust). The local and global path planning problem is further formalized in the following subsection.

\subsection{Path Planning}
Consider a configuration space $\rchi = [0, 1]^d$ of a system, with $d\in\mathbb{N}$ and $d \geq 2$. Some part of the state space contains obstacles, and therefore should be avoided by the planning algorithm. We denote this occupied part as $\rchi_{obs} \subseteq \rchi$, the complement of this set is the free part of the state space $\rchi_{free}=\rchi \setminus \rchi_{obs}$. The initial state $x_{init} \in \rchi$ is the starting location of the vessel, the goal region $\rchi_{goal} \subseteq \rchi$ is the location that the autonomous vessel should reach. We can therefore define every path planning problem by the tuple ($\rchi_{free}, x_{init}, \rchi_{goal}$). A continuous function $\sigma: [0, 1] \rightarrow \mathbb{R}^d$ is a collision free path if $\sigma(t) \in \rchi_{free} \forall t \in [0,1]$. More strictly, it is a feasible path if it is collision free and if $\sigma(0)=x_{init}$ and $\sigma(1) \in \rchi_{goal}$.

Applied to planning in autonomous shipping, our problem is defined as follows:
\begin{itemize}   
    \item $\rchi$ is the normalized Cartesian plane with $d=2$, it contains all the coordinates that the autonomous vessel operates in;
    \item $\rchi_{obs}$ contains regions where the autonomous vessel would collide, such as other vessels, buoys or quay walls;
    \item $x_{init}$ is the initial location of the vessel, $\rchi_{goal}$ is the region which is considered as the destination for the vessel e.g., a certain dock in the port.
\end{itemize}

In autonomous shipping, not every collision-free path is desirable, e.g., some paths are longer then necessary or are very non-smooth, therefore, we need to define optimal path planning. We can define an extra function $c: \Sigma \rightarrow \mathbb{R}_{\geq 0}$ that maps all the paths to a certain cost. This cost function must be designed to make sure that an optimal algorithm actually provides the desired behavior. The goal of optimal path planning is to find the path $\sigma^* \in \Sigma_{feasible}$ where $c(\sigma^*) = \min\{c(\sigma): \sigma \in \Sigma_{feasible}\}$. There exist several 0th order search algorithms that provably converge to the optimal path. A notable example of such a search algorithm is RRT*, proposed by Karaman et al. \cite{rrt_star}. However, these classical planning algorithms are only asymptotically optimal given a certain fixed obstacle set $\rchi_{obs}$, which leaves the question how to act optimally in an environment with dynamic obstacles. It is important to note that in a dynamic environment, updating $\rchi_{free}$ and $x_{init}$ and re-running an optimal planning algorithm every timestep does not guarantee a globally optimal path, since the state of other vessels (which defines $\rchi_{free}$) is assumed to be static in every calculation. On the other hand, methods such as MPC \cite{MPC} optimize over the model of the environment and therefore take the actual optimal path, however, they are limited to their planning horizon and are computationally expensive when nonlinear dynamics are involved. We therefore need a decision-making agent that acts according to current observations in the environment and will have taken the full optimal path $\sigma^*_{dynamic}$ when it has reached its goal region $\rchi_{goal}$. A common theoretical framework to describe problems where sequential actions need to be taken to optimize a long-horizon problem is the Markov Decision Process. The next section defines path planning within that framework.

\subsection{Local Planning in the MDP Framework}

A Markov Decision Process (MDP) is defined by the state space $\mathcal{S}$, the action space $\mathcal{A}$, the reward function $R: \mathcal{S} \times \mathcal{A} \rightarrow \mathbb{R}$ and the transition model $T:\mathcal{S} \times \mathcal{A} \rightarrow \mathcal{S}'$. The state space contains all possible states that the agent can be in. In autonomous shipping the state includes the location of the ownship (OS) (i.e. $x_{current} \in \rchi$), but also other information such as its heading and observed obstacles. Note that every state has exactly one current location $x_{current}$ (a surjective function can be found between them). The action space defines the output that the agent can provide. In this paper, our agent learns to control the desired heading and desired speed, therefore $\mathcal{A} \subset \mathbb{R}^2$. The transition model defines how a certain action $a \in \mathcal{A}$ in a certain state $s \in \mathcal{S}$ leads to a next state $s' \in \mathcal{S}$. A trajectory $\tau$ is a sequence of states and actions in the environment $[(s_0, a_0),(s_1, a_1), ..., (s_{final}, a_{final})]$. Because every state can be mapped to exactly one coordinate in $\rchi$ and every action defines the next state, a trajectory $\tau$ in an MDP defines a path $\sigma$ in the configuration space. The return $G(\tau)$ in an MDP is defined as:
\begin{equation}
    G(\tau) = \sum_{t=0}^H{\gamma^t \cdot R(s_{t}^{\tau}, a_{i}^{\tau})}
\end{equation}, where $H$ denotes the horizon (length of the trajectory) and $\gamma$ is the discount factor, which is used to balance between long-term and short-term effects of a decision. The value $v^\pi(s_t)$ is the expectation of the return starting from a state $s_t$ when acting with policy $\pi$.

A policy $\pi: \mathcal{S} \times \mathcal{A} \rightarrow [0, 1]$ in an MDP defines the probability distribution over actions in a certain state. The optimal policy is defined as follows:
\begin{equation}
    \pi^* = \underset{\pi}{\operatorname{argmax}} \: \mathbb{E}_{\pi} \left[G(\tau|\pi)\right]
\end{equation}

This means that if we set the reward in our MDP, such that $G(\tau) = - c(\sigma)$, the optimal policy $\pi^*$ in the MDP defines the expected optimal path in a dynamic environment $\sigma^*_{dynamic}$. RL is designed to converge to the optimal policy in an MDP, therefore a trained RL agent can produce the optimal path by taking sequential actions, given that the rewards in an episode sum to $-c(\sigma)$.


\section{Methodology} \label{sec:methodology}
The following section describes the approach that has been taken. It first details the simulation that is proposed (the environment of the MDP). Next, we provide details about our state representation and reward formulation respectively. Then, the problem of overfitting is described and an approach to tackle this problem is provided. Finally, a large part of this section is dedicated to MBRL and MuZero \cite{muzero}.

\subsection{Simulation} \label{sec:simlation}
The focus of this paper is not on control but rather on navigational planning, therefore we employ a three degree of freedom (3-DOF) kinematic model, which is an extension of the simulation provided by the MOOS-IVP framework \cite{moos_ivp}. This fast-to-compute model allowed us to speed up the simulation and therefore increase the computational research capacity towards the focus of this paper, i.e. generalization and unstructured observations. Nevertheless, the significant delay between control inputs and physical effects, is still maintained. Also, the vessel is underactuated, since there are three DOFs (surge, sway and yaw) and only two controllers: the rudder (yaw) and the thrust (surge). The model is described as follows:

\begin{itemize}
    \item The current pose vector at time step $t$: $\mathbf{p}_t = [x_t, y_t, \theta_{t}]^T$, where $(x_t, y_t)$ represents the current position and $\theta_t$ is the current heading in degrees.
    \item An informal representation of the time derivative $\frac{\delta\mathbf{p}_t}{\delta t}$ at time step $t$: $\mathbf{v}_t = [s, \dot{\theta}]^T$, where $s$ represents the translational velocity and $\dot{\theta}$ the angular velocity of the vessel.
    \item The control inputs $\mathbf{u}_t = [u_{thrust}, u_{heading}]^T$.
\end{itemize} $\mathbf{v}_t$ is simulated non-linearly as follows: 

\begin{align}
    F_{drag} &=\frac{1}{2} \rho {s^2} c_d A \quad \textrm{(drag equation)} \label{eqn:drag}\\
    a &= \frac{u_{thrust}-F_{drag}}{m} \label{eqn:acceleration} \\
    \mathbf{v}_{t+dt} &= \mathbf{v}_t + 
    \begin{bmatrix}
        a \cdot dt \\
        u_{heading} \cdot T \cdot dt
    \end{bmatrix},
\end{align}  Where Eqn. \ref{eqn:acceleration} calculates the magitude of acceleration on the vessel. $\mathbf{p}_t$ is now updated as:

\begin{equation}
    \mathbf{p}_{t+dt} = \mathbf{p}_t + 
    \begin{bmatrix}
        sin(\theta)(s\cdot dt)\\
        cos(\theta)(s\cdot dt)\\
        \dot{\theta}\cdot dt
    \end{bmatrix}.
\end{equation} Where $T$ is the turn rate (a constant) and $dt$ is the time step. The drag equation (\ref{eqn:drag}) has the following parameters: $\rho$ represents the mass density of the fluid, $A$ is the reference area and $c_d$ is the drag coefficient.

\subsection{State Representation} \label{sec:state}
\begin{figure}
    \centering
    \includegraphics[width=1.0\linewidth]{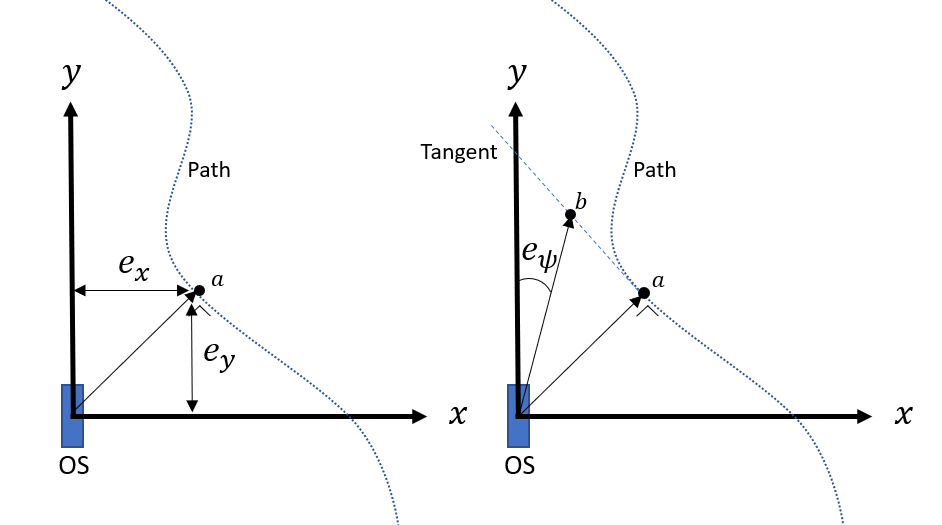}
    \caption{The left part shows: $e_x$ and $e_y$. The right part shows $e_\psi$.}
    \label{fig:path_error}
\end{figure}

Since the sensors on OS can never see the full environment (because of the limited range and shadowing), the RL agent does not take the full state $s_t$ as its input but rather an observation $o_t$ which contains information about the state. The perception of obstacles is computed by performing ray tracing between the OS and all obstacles in the environment. The distance between OS and every intersection with an (dynamic) obstacle is represented as $\mathbf{d}$. All the odometry of OS is contained in the  odometry vector $\mathbf{m} = [\delta, \dot{\delta}, q, s, \theta]$, with $\delta$ the rudder angle, $q$ the distance to the goal and the other scalars as defined in Sect. \ref{sec:simlation}. The errors w.r.t. the global path are contained in $\mathbf{e} = [e_x, e_y, e_\psi]$, this vector contains the line-of-sight (LOS) path following errors as proposed by Fossen et al. \cite{los} and are depicted in Fig. \ref{fig:path_error}. $e_x$ and  $e_y$ are the relative $x$ and $y$ deviations between OS and the closest point on the global path (point $a$ in Fig. \ref{fig:path_error}). $e_\psi$ describes the heading error between the heading of OS and a straight line from OS to a point on the tangent in $a$. That point on the tangent ($b$ in Fig. \ref{fig:path_error}) lies at a fixed distance from $a$, i.e. $dist(a,b)=constant$. Furthermore, the relative coordinates to the ten closest points on the path are added in $\mathbf{z}=[x_0, y_0, ..., x_9, y_9]$.

All these vectors are combined with the previous action of the agent to form the full observation:

\begin{equation}
    \mathbf{o}_{t} = \mathbf{d}_t\mathbin\Vert \mathbf{m}_t \mathbin\Vert \mathbf{e}_t \mathbin\Vert \mathbf{z}_t \mathbin\Vert \mathbf{u}_{t-1}
\end{equation}

Because there is much temporal information in the transition between subsequent time steps, we employ frame stacking, such that the three previous observations are concatenated with the current one using a simple first-in-first-out buffer. This forms the total observation that is presented to the agent:

\begin{equation} \label{eqn:obs}
    \mathbf{o}_{total} = \mathbf{o}_{t}\mathbin\Vert \mathbf{o}_{t-1} \mathbin\Vert \mathbf{o}_{t-2} \mathbin\Vert \mathbf{o}_{t-3}
\end{equation}



\subsection{Reward Shaping} \label{sec:reward}

The reward should motivate the agent to stick to the global path as best as possible while there is no risk for a collision. The agent should avoid dynamic obstacles at all cost while navigating along the path to the goal as a second priority. Therefore, we define the total reward function by balancing the encouraging reward $R(s_t)$ that incentivizes the agent to follow the path closely and the constraint cost $C(s_t)$ that indicates if a collision happens. The reward function is designed as follows: 

\begin{align}
    r_{total} &= C_r \cdot R(s_t) + C_c \cdot C(s_t) \\
    R(s_t) &= R_{path}(s_t) + R_{goal}(s_t) 
\end{align} with:

\begin{align}
     C(s_t) &= \mathbbm{1}_{\rchi_{obs}}(x_t, y_t) \\
    R_{goal}(s_t) &= \mathbbm{1}_{\rchi_{goal}}(x_t, y_t) * 1000 \\
    R_{path}(s_t) &= -\left\lVert \mathbf{e}_{t} \right\rVert_{1} 
\end{align}

Where $C(s_t)$ discourages the agent for hitting an obstacle and we followed \cite{los_planning} in defining the path-following reward $R_{path}(s_t)$. Finally, $R_{goal}(s_t)$ is an extra bonus for the agent if it completes an episode by reaching the goal. In our work, we choose $C_r = 1$ and $C_c = -1000$.



\subsection{Domain Randomization} \label{sec:randomization}
\begin{figure}
    \centering
    \includegraphics[width=0.75\linewidth]{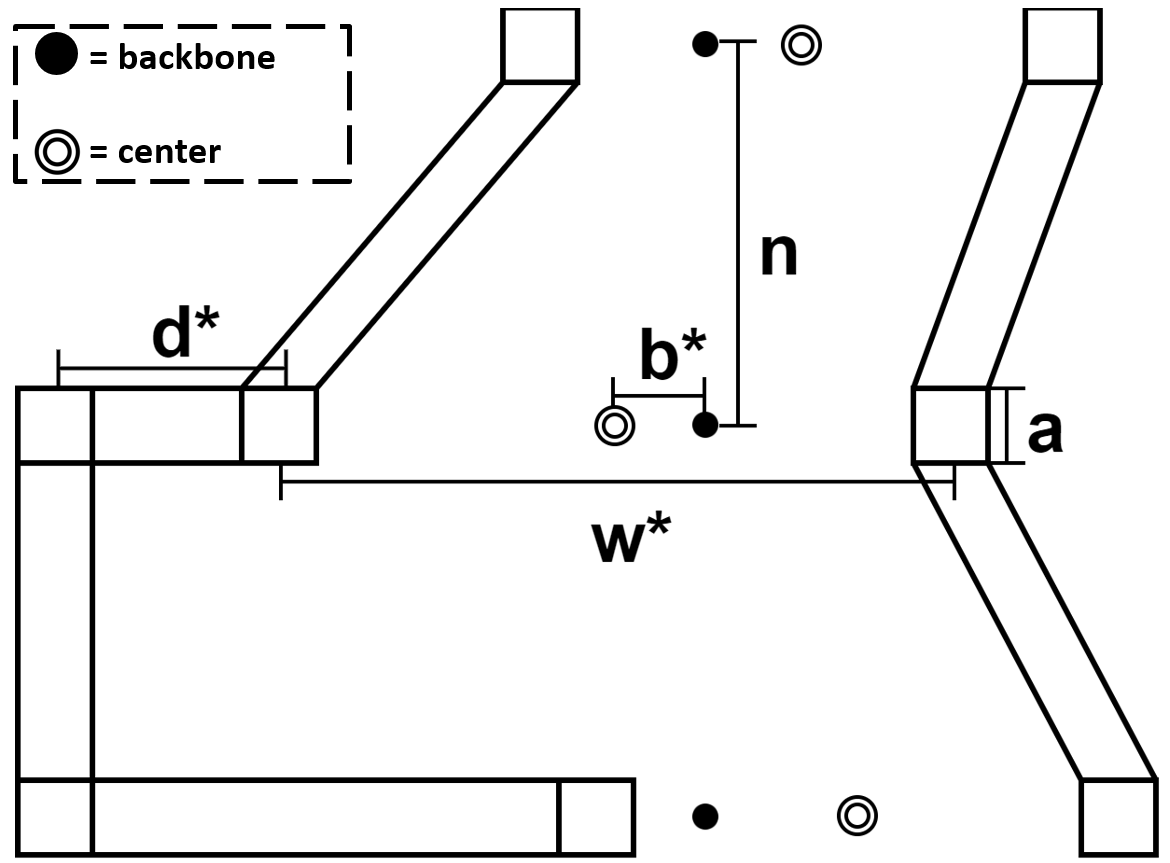}
    \caption{Randomization of the Shipping Environment.}
    \label{fig:randomization}
\end{figure}

RL policies suffer from distributional shift: they do not work outside of the configuration space that they were trained in. Formally, this happens because samples from the state space $\mathcal{S}_{train}$ of that port are not i.i.d. samples from another port with state space $\mathcal{S}_{test}$. We therefore need to train our agent with representative samples of a distribution $\mathcal{S}_{train} \supseteq \mathcal{S}_{test}$. The case where $\mathcal{S}_{train} = \mathcal{S}_{test}$ means that our agent only learns to navigate in a single port scenario. Since the goal of this paper is to provide a general navigation algorithm, we make sure that $\lvert \mathcal{S}_{train} \rvert$ is many times larger than $\lvert \mathcal{S}_{test} \rvert$, forcing the agent to generalize over a large amount of possible port configurations. To provide a sufficiently large $\mathcal{S}_{train}$ we employ domain randomization \cite{domain_randomization} in our simulator by randomizing the following elements (see Fig. \ref{fig:randomization}), based on the docks found in the Port of Antwerp:
\begin{itemize}
    \item The length of a dock: $d^* \sim \mathpalette{U}(300, 900)$,
    \item Center of the waterway, with reference to the backbone: $b^* \sim \mathpalette{U}(-600, 600)$,
    \item The width of the waterway: $w^* \sim \ \mathpalette{U}(300, 900)$.
\end{itemize} Furthermore, we randomize the starting location of every other vessel in the environment and their goal regions, this makes sure that the behavior of simulated vessels can not be memorized by the agent, but needs to be predicted by their current movements. The goal position of OS is randomized as well. Since every path planning problem is defined by the tuple ($\rchi_{free}, x_{init}, \rchi_{goal}$) and the global path is part of the state space, randomizing the goal region $\rchi_{goal}$ and the starting location ${x_{init}}$ also randomizes the state space.

\subsection{Model-based Reinforcement Learning and MuZero} \label{sec:muzero}

This section first describes the key idea behind model-based RL. Next we provide an overview of the model-based RL algorithm that is used in this paper: MuZero \cite{muzero}. Finally, we provide the objective function that is minimized during training. 

\subsubsection{Model-based Reinforcement Learning}
As the name suggests, a key part of MBRL is the transition model $T: \mathcal{S} \times \mathcal{A} \rightarrow \mathcal{S}'$. Although some works provide that function to the agent, we make no such assumption and therefore learn $T$ from environment experience. The advantage of having this transition function is that these predictions can be used in a planning algorithm to improve the long term reasoning capability of an RL agent. Since $T$ only makes predictions for a single step instead of learning a quantity over the full horizon, this function is often easier to learn. The agent can then use imagination (planning) over that model to learn other functions such as the value or the policy, improving the data efficiency. To clarify these abstract concepts, the model-based algorithm that was used in this paper is explained thoroughly in the next section.

\subsubsection{Operation}
\begin{figure}
    \centering
    \includegraphics[width=.7\linewidth]{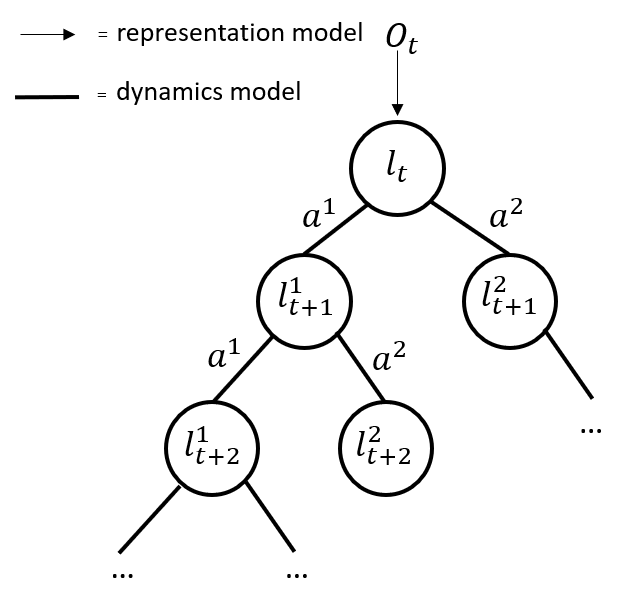}
    \caption{The tree search of MuZero.}
    \label{fig:muzero}
\end{figure}

MuZero \cite{muzero} is a state-of-the-art MBRL algorithm that combines learning and planning. It employs three learned functions (represented by neural networks) during its search: 

\begin{itemize}
    \item the representation function $h_{\phi}$, which maps an observation $\mathbf{o}_{t}$ to a latent state $\mathbf{l}_{t}$,
    \item the transition function $g_{\phi}$ that predicts $r_{t+1}$ and $\mathbf{l}_{t+1}$ from $\mathbf{l}_{t}$ and $a_{t-1}$,
    \item the prediction function $f_{\phi}$ takes a latent state $\mathbf{l}_{t}$ to predict its value $v_t$ and a predicted policy $\pi'$.
\end{itemize} all these functions depend on one combined set of parameters $\phi$, which are the weights and biases of the neural networks.


To understand how these functions define the decision making process, lets first assume they are perfectly trained already. First, the agent takes an observation from the environment at time step $t$, as defined by Eqn. \ref{eqn:obs}. This observation is provided as the input to $h_\phi$, which outputs a latent state: $\mathbf{l}_{t} = h_\phi(\mathbf{o}_{t})$. This latent state is a lower-dimensional representation of the actual observation. The agent is fully allowed to represent the observation in any way it wants, as long as this representation helps the agent to accurately predict $r_{t+1}$, $v_t$ and $\pi'$ in the following steps. Concretely, our representation model takes a \mbox{$dim(\mathbf{o}_{total})$-dimensional} input and outputs a 32-dimensional latent state (32 floating point numbers).

Second, this latent state is used as the root node in a planning algorithm called Monte Carlo Tree Search (MCTS) \cite{mcts}, which has the goal of finding the best possible action at the root, based on the sum of rewards that are predicted throughout the search. This is the purpose of $g_{\phi}$: for every action that is explored in a node it can predict the next node and the corresponding reward: $r_{t+(n+1)}, \mathbf{l}_{t+(n+1)} = g_\phi(a_{t+n}, \mathbf{l}_{t+n})$ (see Fig. \ref{fig:muzero}). Note that this function is the transition function $T$, only represented in latent space instead of observation space. A search could now be performed, which maximizes the reward over every possible trajectory from the root and then takes that action. However, this quickly becomes computationally intractable. Therefore we need to limit the depth and the width of the tree with estimations instead of the actual search values. Since the value is an estimation of the expected return from a certain (latent) state, we can search until the computational budget is exhausted, after which we perform an estimation of how much reward will still be collected from a specific leaf node: $v_{t+d} = f_{\phi}(\mathbf{l}_{t+d})$, with $d$ the depth of the leaf node. The predicted rewards on a certain search trajectory summed with this value estimation in the root node is now the best estimate of what return an action in the root node will deliver. 

The following three steps compile the descision making process:

\begin{enumerate}
    \item Encode the current observation with $h_\phi$,
    \item Predict an optimal policy $\pi'$ using $f_\phi$,
    \item Perform a MCTS with $g_\phi$ as the transition model. The search is guided by $\pi'$ as a prior belief, but improves this policy using the predicted rewards from $g_\phi$ and a value prediction from $f_\phi$ in the leaf nodes. This improved policy is $\bar\pi$,
    \item Sample an action accoring to the distribution $\bar\pi$.
\end{enumerate}

We can summarize this process (rather informally) as:

\begin{align}
    v^{tree}_t, \bar\pi &= MCTS(\mathbf{o}_t, g_\phi, h_\phi, f_\phi) \\
    a_t &\sim \bar\pi
\end{align} where $v^{tree}_t$ is the tree search value, which is an updated value estimate of the root node, based on the predicted rewards and leaf-node values during the search.

The exact methodology of exploring during the search and combining the predicted rewards, values and predicted policy to form $\bar\pi$ and $v^{tree}_t$ is left out of this work due to space constraints. It is well explained in \cite{muzero}, updated with some changes that were proposed in \cite{pozero}. The adaptions from \cite{pozero} were a key element to reach stable training performance and should not be overlooked by anyone reproducing this work. 

After every action that is performed in the environment, the tuple $(a_t, \mathbf{o}_t, v^{tree}_t, \bar\pi)$ is stored in a buffer $\mathcal{D}$ to be used for training the neural networks. That training process is described next.

\subsubsection{Learning} \label{sec:mz_learning}

To learn, MuZero optimizes three objectives. The first objective is the n-step bootstrapping objective, which is an extension from standard TD-learning \cite{sutton_barto}. The goal is to minimize the expected mean squared error between the predicted value $V_\phi(\mathbf{o}_t)$ at time step $t$ and the actual return for $n$ steps summed with the tree search value at time step $t+n$. This means minimizing the following objective:

\begin{equation} \label{value_objective}
    J_V(\phi) = \mathbb{E}_{(\mathbf{o}_t, r_t, v^{tree}_t) \sim \mathcal{D}} \left[ (V_\phi(\mathbf{o}_t) - \hat{V}(\mathbf{o}_t))^2 \right],
\end{equation}

with 
\begin{equation} 
    \hat{V}(\mathbf{o}_t) = \sum_{k=0}^{n-1} \gamma^k r(\mathbf{o}_{t+k}) +\gamma^n v^{tree}_{t+n}.
\end{equation}

The second objective is to minimize the mean squared error between the predicted reward $R_\phi(\mathbf{o}_t)$ at a certain time step with the actual received reward. This leads to minimizing the following objective:
\begin{equation} \label{reward_objective}
    J_R(\phi) = \mathbb{E}_{(\mathbf{o}_t, r_t) \sim \mathcal{D}} \left[(R_\phi(\mathbf{o}_t) - r_t)^2 \right].
\end{equation}

The third and final objective is to minimize the KL-divergence between the predicted policy $\pi'(.|\mathbf{o}_t)$ and the improved search policy $\bar{\pi}(.|\mathbf{o}_t)$:
\begin{equation} \label{policy_objective}
    J_\pi(\phi) = \mathbb{E}_{\mathbf{o}_t \sim \mathcal{D}} \left[ \mathcal{KL}(\bar{\pi}(.|\mathbf{o}_t), \pi'(.|\mathbf{o}_t)) \right].
\end{equation}

As noticeable from equations \ref{value_objective} trough \ref{policy_objective}, all objectives are parameterized by one set of parameters $\phi$. This means that we can jointly optimize these objectives by minimizing the total MuZero objective: 

\begin{equation} \label{eqn:total_objective}
J_{total}(\phi) = J_V(\phi) + J_R(\phi) + J_\pi(\phi).
\end{equation} This minimization can be done by a gradient based method, in this work, we follow \cite{pozero} in employing the Adam \cite{adam} optimizer.

This objective formulation leads to the natural question of how the transition model $T(\cdot)$ is learned, since it does not seem to appear in the total MuZero objective. The answer lays in the fact that predictions are not performed directly on an observation. Instead, an observation is forwarded trough the representation model, after which it is forwarded trough the transition model for $k$ times. This $k$-step rollout is then used to make all the predictions that are used to compute Eqn. \ref{eqn:total_objective}. The chain rule of derivatives therefore makes sure that $h_{\phi}$ and $g_{\phi}$ are also optimized by minimizing the total objective. Note again that these two functions together are the total transition function $T(.)$. Since $g_{\phi}$ takes its own output (together with a recorded action) as the next input, this is a recurrent neural network \cite{rnn}. For an unroll of depth $k$, this process is summarized as:

\begin{equation}
    V_{\phi}, \pi_{\phi} = (f_{\phi} \circ \underbrace{g^{a_{t+k}}_{\phi} \circ ... \circ g^{a_t}_{\phi}}_k \circ h_{\phi})(\mathbf{o}_t) 
\end{equation}

This has the result that that the representation - and dynamics model only optimize their approximation of $T$ with the goal of being as useful as possible to predict a reward, value and policy. All aspects of the environment which are irrelevant to optimizing the performance of the agent are therefore ignored by the model.

\section{Experimental setup}

Our agent is trained and tested in a custom simulator that is based on the MOOS-IVP \cite{moos_ivp} framework. The agent is the MuZero algorithm as described in Sect. \ref{sec:muzero}. It provides actions in a 2D action space $\mathcal{A}=\mathcal{H} \times \mathcal{P}$, where $\mathcal{H}$ is a set that contains 9 discrete, evenly spaced values between -30 and 30 which represent the desired change in heading for OS (in degrees). $\mathcal{P}=\{\textrm{dead slow}, \textrm{half}, \textrm{full}\}$ represents 3 discrete desired speed values for the vessel (1m/s, 2.5m/s, 5m/s). The desired heading and speed from the OS are passed to a proportional-integral-derivative (PID) controller which sets the values of $u_{thrust}$ and $u_{heading}$. These signals control the motion of the vessel as described in Sect. \ref{sec:simlation}. On every reset of our agent (after a collision or when the OS reaches its goal), six other vessels select a random start location and goal region in one of the docks, after which they calculate a path using RRT* \cite{rrt_star}. The default waypoint behavior from MOOS-IVP makes sure that the vessels follow that global path to their goal. Every 10th reset, a new environment is generated with the methodology described by \ref{sec:randomization}. This generation algorithm does not only provide static obstacles (the docks and buoys), but also a valid starting point and goal regions for both the OS and the other vessels. To train the RL agent, we implemented a performant, distributed version of which allowed us to run 25 workers in parallel which asynchronously collect data in separate simulations.   

\section{Results and discussion} \label{sec:results}
This section describes the results of our research. First the qualitative results validate our research by showing the operation of our agent in a randomly selected scenario. Second, the quantitative results are analyzed to evaluate the performance of the agent over many scenarios. Third, we provide a quantitative comparison between our method, the Dynamic Window Approach (DWA) and a model-free RL method. 



\begin{figure}
    \centering
    \includegraphics[width=.62\linewidth]{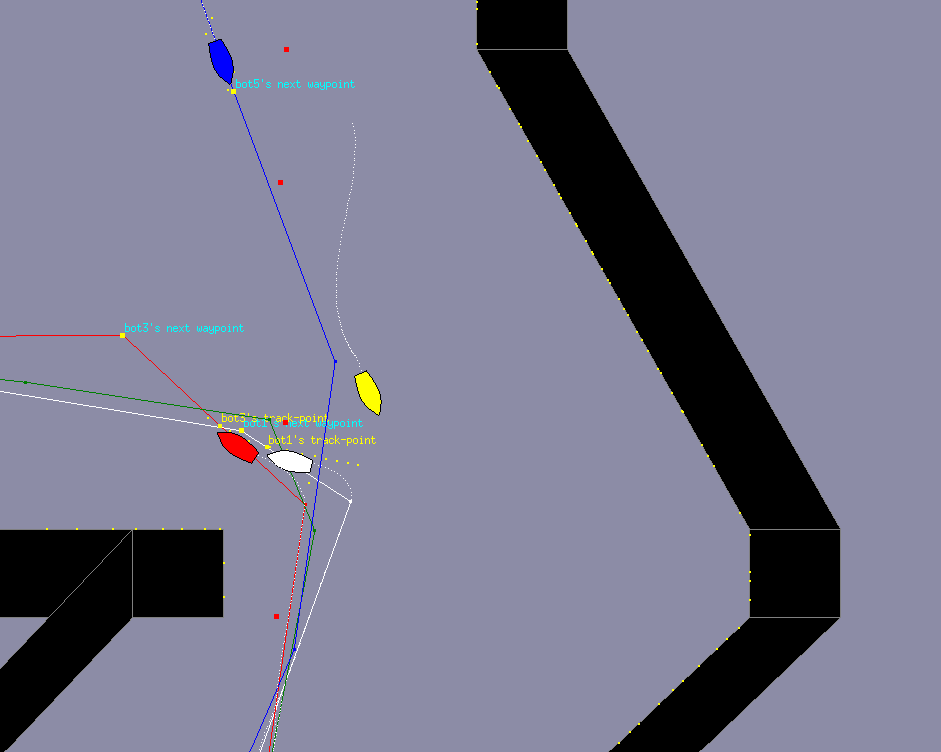}
    \caption{OS evades other vessels.}
    \label{fig:wijkt_uit}
\end{figure}

\begin{figure}
    \centering
    \includegraphics[width=.5\linewidth]{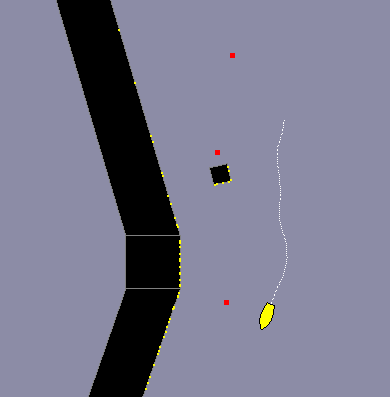}
    \caption{OS evades a buoy.}
    \label{fig:boei}
\end{figure}


\begin{figure*} [t]
\begin{subfigure}[t]{0.5\textwidth} 
\vbox{
\vspace*{1.7em}
\centering{
  \includegraphics[width=0.6\textwidth]{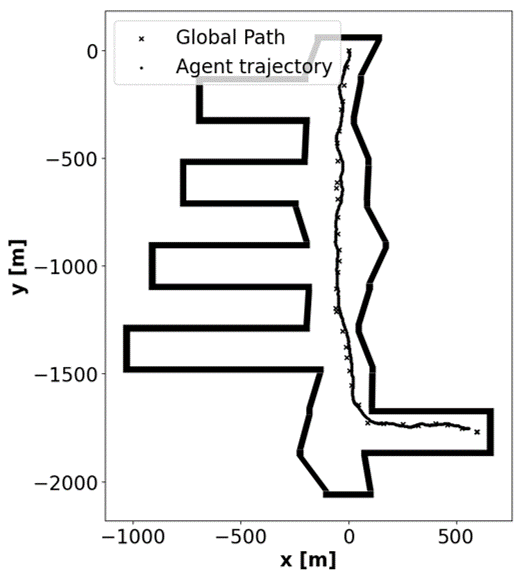}
}%
}%
\subcaption{OS encounters no dynamic obstacles}
\end{subfigure}%
\begin{subfigure}[t]{0.5\textwidth}
\centering{%
\includegraphics[width=0.51\textwidth]{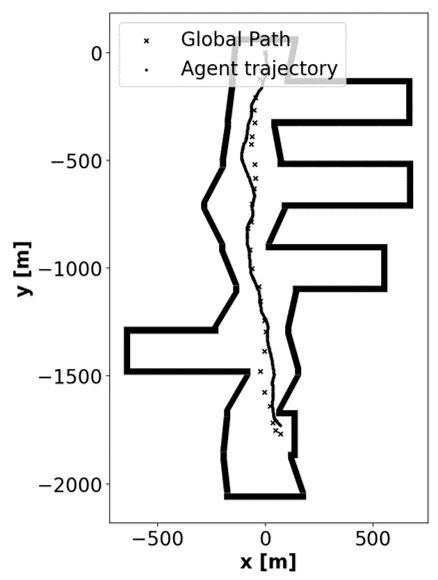}
}%
\subcaption{OS encounters dynamic obstacles}
\end{subfigure}
\caption{The recorded trajectory of OS in two randomly selected scenarios.} \label{fig:examples}
\end{figure*}


\begin{figure*}
    \centering
    \includegraphics[width=.8\textwidth]{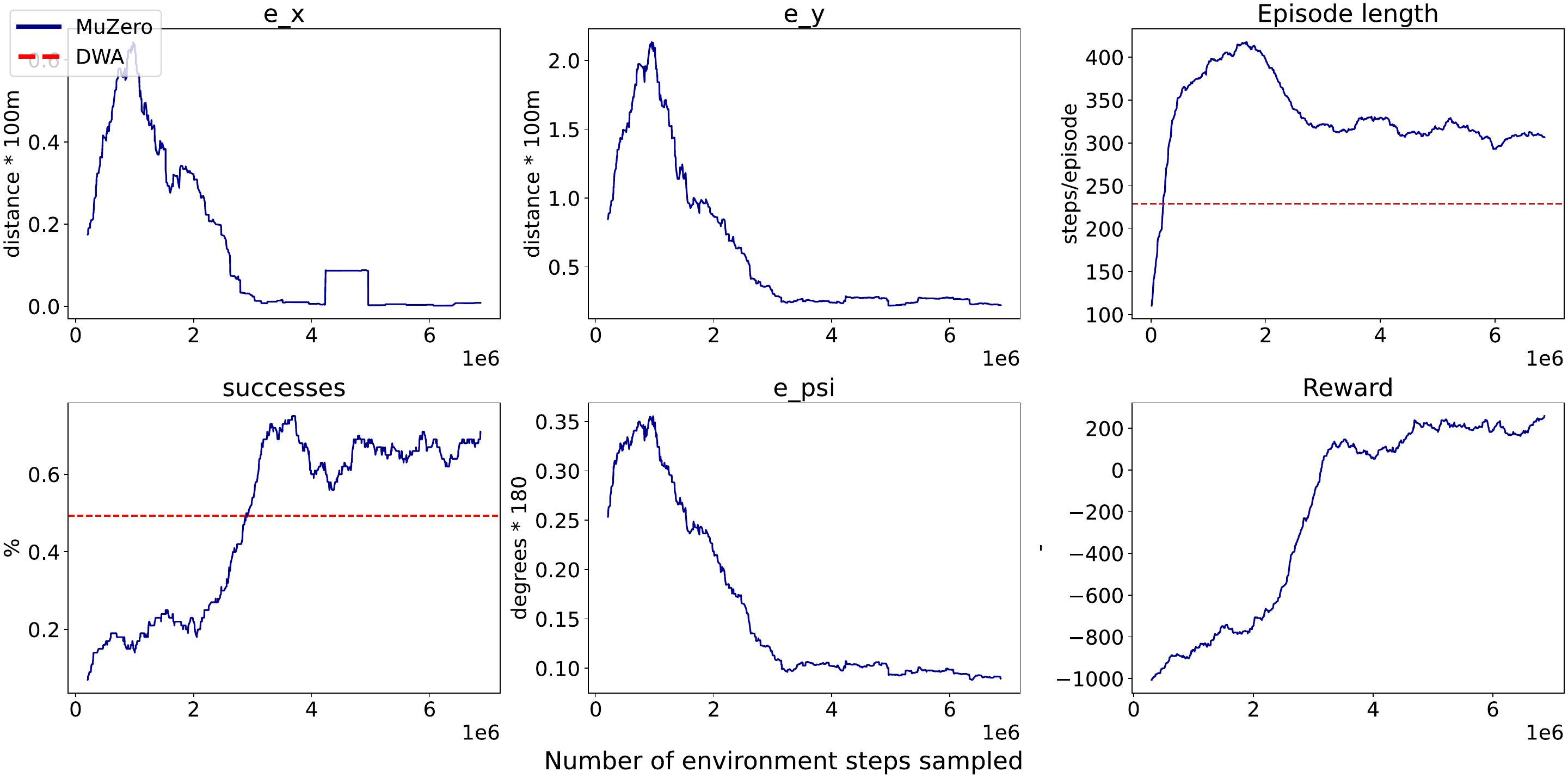}
    \caption{Metrics logged during the training process. e_x, e_y, e_psi: path following errors (Sect. \ref{sec:state}); a success represents a full collision-free episode (start to goal); reward as in Sect. \ref{sec:reward}; episode length: steps (=seconds) per episode. A running average filter (window size 100) was used. DWA baseline included for relevant metrics.}
    \label{fig:training}
\end{figure*}

\subsection{Qualitative Results}

Fig. \ref{fig:examples} shows the path of the OS in two random scenarios. The agent has never seen these scenarios during training. Subfigure (a) shows how the OS remains on the global path from start to finish until it reaches its end goal (which is also randomly generated). The path is about 2.7km long. Subfigure (b) shows the OS diverging from its path around point (0, -500) and (0, -1500) to evade dynamic obstacles. It is visible that OS recovers from the evasion with a smooth trajectory back to the global path. This "smoothness" is incentivized by the LOS path following reward (Sec. \ref{sec:reward}). Furthermore, Fig. \ref{fig:boei} and \ref{fig:wijkt_uit} show two screenshots from the proposed simulator. OS is visible as a yellow vessel, other vessels and their paths are visible in different colors. The global path is shown as small red squares. Fig. \ref{fig:wijkt_uit} shows a scenario where OS has to diverge from its global path to evade the red and white vessels. It can be seen that OS successfully performs an evasive maneuver after detecting the other vessels using its ranging sensors. On the other hand, Fig. \ref{fig:boei} shows an evasive maneuver after OS detects a buoy in its field of view. In the screenshot, the OS has already evaded the buoy and is heading back to the global path.

\subsection{Quantitative Results}

Fig. \ref{fig:training} shows the evolution of the path following and collision avoidance performance during the training process. It is visible that the MuZero agent achieves a 70\% success rate after 3 million environment steps or about 10 thousand episodes in the environment. This success rate defines the fraction of episodes that fully traverses the global path without a collision. A fully trained agent has an average episode length of 300s, this means that the agent is able to sail safely for a period of 15 minutes on average, before encountering a safety violation. Furthermore, all the subfigures in Fig. \ref{fig:training} confirm that the proposed reward correctly prioritizes collision avoidance over path following. First, the episode length immediately increases from the start, meaning that the first priority of the agent is to learn to not collide. Second, at this stage, the path errors also sharply increase, again confirming that the agent does not prioritize path following. After the intial phase (around time step 1M), the agent starts to learn that following the path is a good guidance for reaching the goal, hence it can be seen that the success rate sharply increases while the path errors decrease. The reward also sharply increases at this point. This is expected, since every success provides 2000 reward in total to the agent: +1000 for reaching the goal and not receiving the -1000 since no collision happened. Therefore, Fig. \ref{fig:training} strongly indicates that the proposed reward is well-designed, it correctly incentivizes the agent to follow the global path while avoiding collisions.

\subsection{Comparison with the Dynamic Window Approach}
The success rate and episode length in Fig. \ref{fig:training} compare our approach with the Dynamic Window Approach (DWA), which is a commonly used collision avoidance algorithm in autonomous shipping \cite{dwa_shipping} \cite{dwa_shipping2}. The provided DWA baseline is an average of 100 subsequent episodes in the shipping environment. It is visible that the proposed methodology significantly outperforms DWA with a success rate of 70\% compared to 55\% of DWA. This supports our theoretical observations in Sec. \ref{sec:formulation}, since DWA cannot predict the motion of other vessels, it is often trapped in irrecoverable scenarios. The other metrics in Fig. \ref{fig:training} do not show a baseline, since DWA does not employ the LOS approach \cite{los}.

\subsection{Comparison with Model-Free RL}







\begin{table*}
\caption[Table]{A comparison between a state-of the art model-free agent (PPO) and MuZero.} \label{tab:ppo_mz} 
\centering{%
\begin{tabular}{|llllllllllll|}
\toprule

\emph{time step} & 0 & 500k & 1M & 1.5M & 2M & 2.5M & 3M & 3.5M & 4M & 4.5M & 5M \\
\midrule
\textbf{MuZero} & -1100 & -900 & -800 & -760 & -800 & -200 & 0 & 150 & 75 & 210 & 220 \\ 
\textbf{PPO} & -1100 & -680 & -480 & -470 & -760 & -940 & -270 & -570 & -780 & -1045 & -1056\\
\bottomrule
\end{tabular}
}
\end{table*}

Table \ref{tab:ppo_mz} shows the training progress of MuZero \cite{muzero} and PPO \cite{ppo} respectively. PPO is a model-free reinforcement learning algorithm that has also been used in previous work within the context of autonomous shipping \cite{shipping_rl1}. Although PPO quickly finds a suboptimal policy within 500 thousand time steps, the performance afterwards oscillates and no good policy is found within the data budget. MuZero, however, succeeds in finding a much better policy before stagnating. We suspect that this is because MBRL can explore in two ways: the traditional way where the agent performs exploratory actions in the environment, but also by imagining many possibilities by using the learned transition model during the search phase. This concept is known as two-phase exploration \cite{mbrl_survey}.



\section{Conclusion}
This paper introduces model-based reinforcement learning to the field of autonomous shipping. Results show that our approach is able to provide generalized navigation of autonomous vessels in a wide variety of port scenarios. This generalization is achieved by employing domain randomization. It is shown that our path planning methodology provides safer behavior than the dynamic window approach and we show the advantage of model-based reinforcement learning over a benchmark model-free reinforcement learning algorithm. Because our proposed algorithm uses ranging sensor observations without any ground-truth knowledge of other vessels, this paper's approach is a step toward autonomous navigation in complex inland - and port scenarios.         


\section*{Acknowledgments}
This work is conducted within the Horizon 2020 PIONEERS project, funded by the European Commission under agreement No. 101037564.



\bibliographystyle{asmeconf}  
\bibliography{bibliography}

\begin{thebibliography}{10}
\newcommand{\enquote}[1]{``#1''}
\providecommand{\url}[1]{\texttt{#1}}
\providecommand{\urlprefix}{URL }
\expandafter\ifx\csname urlstyle\endcsname\relax
  \providecommand{\doi}[1]{DOI \discretionary{}{}{}#1}\else
  \providecommand{\doi}{DOI \discretionary{}{}{}\begingroup
  \urlstyle{rm}\Url}\fi
\providecommand{\eprint}[2][]{\urlprefix\url{#1#2}}

\bibitem{eucomm}
European~Commission, Directorate-General for~Mobility and Transport.
\newblock \textit{Assessment of potential of maritime and inland ports and
  inland waterways and of related policy measures, including industrial policy
  measures : final report}.
\newblock Publications Office (2020).
\newblock \urlprefix\url{https://data.europa.eu/doi/10.2832/03796}.

\bibitem{safetyshipping}
Wr{\'o}bel, Krzysztof, Montewka, Jakub and Kujala, Pentti.
\newblock \enquote{Towards the assessment of potential impact of unmanned
  vessels on maritime transportation safety.}
\newblock \textit{Reliability Engineering \& System Safety} Vol. 165 (2017):
  pp. 155--169.
\newblock \doi{10.1016/j.ress.2017.03.029}.

\bibitem{shippingreview}
Ziajka-Pozna{\'n}ska, Ewelina and Montewka, Jakub.
\newblock \enquote{Costs and benefits of autonomous shipping—a literature
  review.}
\newblock \textit{Applied Sciences} Vol.~11 No.~10 (2021): p. 4553.
\newblock \doi{10.3390/app11104553}.

\bibitem{shippingvision}
Chen, Zhijun, Chen, Depeng, Zhang, Yishi, Cheng, Xiaozhao, Zhang, Mingyang and
  Wu, Chaozhong.
\newblock \enquote{Deep learning for autonomous ship-oriented small ship
  detection.}
\newblock \textit{Safety Science} Vol. 130 (2020): p. 104812.
\newblock \doi{10.1016/j.ssci.2020.104812}.

\bibitem{shipping_rl1}
Chun, Do-Hyun, Roh, Myung-Il, Lee, Hye-Won, Ha, Jisang and Yu, Donghun.
\newblock \enquote{Deep reinforcement learning-based collision avoidance for an
  autonomous ship.}
\newblock \textit{Ocean Engineering} Vol. 234 (2021): p. 109216.
\newblock \doi{10.1016/j.oceaneng.2021.109216}.

\bibitem{shipping_rl2}
Jiang, Lingling, An, Lanxuan, Zhang, Xinyu, Wang, Chengbo and Wang, Xinjian.
\newblock \enquote{A human-like collision avoidance method for autonomous ship
  with attention-based deep reinforcement learning.}
\newblock \textit{Ocean Engineering} Vol. 264 (2022): p. 112378.

\bibitem{sutton_barto}
Sutton, Richard~S and Barto, Andrew~G.
\newblock \textit{Reinforcement learning: An introduction}.
\newblock MIT press (2018).

\bibitem{dqn}
van Hasselt, Hado, Guez, Arthur and Silver, David.
\newblock \enquote{Deep Reinforcement Learning with Double Q-Learning.}
\newblock \textit{Proceedings of the AAAI Conference on Artificial
  Intelligence} Vol.~30 No.~1.
\newblock \doi{10.1609/aaai.v30i1.10295}.

\bibitem{ppo}
Schulman, John, Wolski, Filip, Dhariwal, Prafulla, Radford, Alec and Klimov,
  Oleg.
\newblock \enquote{Proximal policy optimization algorithms.}
\newblock \textit{arXiv preprint arXiv:1707.06347}
  \doi{10.48550/arXiv.1707.06347}.

\bibitem{mbrl_survey}
Moerland, Thomas~M, Broekens, Joost and Jonker, Catholijn~M.
\newblock \enquote{Model-based reinforcement learning: A survey.}
\newblock \textit{arXiv preprint arXiv:2006.16712}
  \doi{10.48550/arXiv.2006.16712}.

\bibitem{muzero}
Schrittwieser, Julian, Antonoglou, Ioannis, Hubert, Thomas, Simonyan, Karen,
  Sifre, Laurent, Schmitt, Simon, Guez, Arthur, Lockhart, Edward, Hassabis,
  Demis, Graepel, Thore et~al.
\newblock \enquote{Mastering atari, go, chess and shogi by planning with a
  learned model.}
\newblock \textit{Nature} Vol. 588 No. 7839 (2020): pp. 604--609.
\newblock \doi{10.1038/s41586-020-03051-4}.

\bibitem{mb_mpo}
Clavera, Ignasi, Rothfuss, Jonas, Schulman, John, Fujita, Yasuhiro, Asfour,
  Tamim and Abbeel, Pieter.
\newblock \enquote{Model-based reinforcement learning via meta-policy
  optimization.}
\newblock \textit{Conference on Robot Learning}: pp. 617--629. 2018. PMLR.
\newblock \urlprefix\url{https://proceedings.mlr.press/v87/clavera18a.html}.

\bibitem{dwa_shipping}
Kim, Hyo-Gon, Yun, Sung-Jo, Choi, Young-Ho, Ryu, Jae-Kwan and Suh, Jin-Ho.
\newblock \enquote{Collision Avoidance Algorithm Based on COLREGs for Unmanned
  Surface Vehicle.}
\newblock \textit{Journal of Marine Science and Engineering} Vol.~9 No.~8.
\newblock \doi{10.3390/jmse9080863}.

\bibitem{fm_rl}
Wang, Shuwu, Yan, Xinping, Ma, Feng, Wu, Peng and Liu, Yuanchang.
\newblock \enquote{A novel path following approach for autonomous ships based
  on fast marching method and deep reinforcement learning.}
\newblock \textit{Ocean Engineering} Vol. 257 (2022): p. 111495.
\newblock \doi{10.1016/j.oceaneng.2022.111495}.

\bibitem{fast_marching}
Sethian, James~A.
\newblock \enquote{A fast marching level set method for monotonically advancing
  fronts.}
\newblock \textit{Proceedings of the National Academy of Sciences} Vol.~93
  No.~4 (1996): pp. 1591--1595.
\newblock \doi{10.1073/pnas.93.4.1591}.

\bibitem{los}
Fossen, Thor~I., Breivik, Morten and Skjetne, Roger.
\newblock \enquote{Line-of-sight path following of underactuated marine craft.}
\newblock \textit{IFAC Proceedings Volumes} Vol.~36 No.~21 (2003): pp.
  211--216.
\newblock \doi{10.1016/S1474-6670(17)37809-6}.
\newblock 6th IFAC Conference on Manoeuvring and Control of Marine Craft (MCMC
  2003), Girona, Spain, 17-19 September, 1997.

\bibitem{los_planning}
Peng, Zhouhua, Liu, Enrong, Pan, Chao, Wang, Haoliang, Wang, Dan and Liu, Lu.
\newblock \enquote{Model-based Deep Reinforcement Learning for Data-driven
  Motion Control of an Under-actuated Unmanned Surface Vehicle: Path Following
  and Trajectory Tracking.}
\newblock \textit{Journal of the Franklin Institute}
  \doi{10.1016/j.jfranklin.2022.10.020}.

\bibitem{nmpc}
Abdelaal, Mohamed, Fränzle, Martin and Hahn, Axel.
\newblock \enquote{Nonlinear Model Predictive Control for trajectory tracking
  and collision avoidance of underactuated vessels with disturbances.}
\newblock \textit{Ocean Engineering} Vol. 160 (2018): pp. 168--180.
\newblock \doi{10.1016/j.oceaneng.2018.04.026}.

\bibitem{colav_attention}
Jiang, Lingling, An, Lanxuan, Zhang, Xinyu, Wang, Chengbo and Wang, Xinjian.
\newblock \enquote{A human-like collision avoidance method for autonomous ship
  with attention-based deep reinforcement learning.}
\newblock \textit{Ocean Engineering} Vol. 264 (2022): p. 112378.
\newblock \doi{10.1016/j.oceaneng.2022.112378}.

\bibitem{smart_waterways}
Vanneste, Astrid, Vanneste, Simon, Vasseur, Olivier, Janssens, Robin, Billast,
  Mattias, Anwar, Ali, Mets, Kevin, De~Schepper, Tom, Mercelis, Siegfried and
  Hellinckx, Peter.
\newblock \enquote{Safety Aware Autonomous Path Planning Using Model Predictive
  Reinforcement Learning for Inland Waterways.}
\newblock \textit{IECON 2022 – 48th Annual Conference of the IEEE Industrial
  Electronics Society}: pp. 1--6. 2022.
\newblock \doi{10.1109/IECON49645.2022.9968678}.

\bibitem{apf_rl}
Zhang, Xinyu, Wang, Chengbo, Liu, Yuanchang and Chen, Xiang.
\newblock \enquote{Decision-Making for the Autonomous Navigation of Maritime
  Autonomous Surface Ships Based on Scene Division and Deep Reinforcement
  Learning.}
\newblock \textit{Sensors} Vol.~19 No.~18.
\newblock \doi{10.3390/s19184055}.

\bibitem{cnn_smallships}
Chen, Zhijun, Chen, Depeng, Zhang, Yishi, Cheng, Xiaozhao, Zhang, Mingyang and
  Wu, Chaozhong.
\newblock \enquote{Deep learning for autonomous ship-oriented small ship
  detection.}
\newblock \textit{Safety Science} Vol. 130 (2020): p. 104812.
\newblock \doi{10.1016/j.ssci.2020.104812}.

\bibitem{rrt_star}
Karaman, Sertac and Frazzoli, Emilio.
\newblock \enquote{Sampling-based algorithms for optimal motion planning.}
\newblock \textit{The international journal of robotics research} Vol.~30 No.~7
  (2011): pp. 846--894.
\newblock \doi{10.1177/0278364911406761}.

\bibitem{MPC}
Gr{\"u}ne, Lars and Pannek, J{\"u}rgen.
\newblock \enquote{Nonlinear model predictive control.}
\newblock \textit{Nonlinear model predictive control}.
\newblock Springer (2017): pp. 45--69.
\newblock \doi{10.1007/978-3-642-01094-1}.

\bibitem{moos_ivp}
Benjamin, Michael~R, Leonard, John~J, Schmidt, Henrik and Newman, Paul~M.
\newblock \enquote{An overview of moos-ivp and a brief users guide to the ivp
  helm autonomy software.} .

\bibitem{domain_randomization}
Tobin, Josh, Fong, Rachel, Ray, Alex, Schneider, Jonas, Zaremba, Wojciech and
  Abbeel, Pieter.
\newblock \enquote{Domain randomization for transferring deep neural networks
  from simulation to the real world.}
\newblock \textit{2017 IEEE/RSJ international conference on intelligent robots
  and systems (IROS)}: pp. 23--30. 2017. IEEE.
\newblock \doi{10.1109/IROS.2017.8202133}.

\bibitem{mcts}
Chaslot, Guillaume, Bakkes, Sander, Szita, Istvan and Spronck, Pieter.
\newblock \enquote{Monte-carlo tree search: A new framework for game ai.}
\newblock \textit{Proceedings of the AAAI Conference on Artificial Intelligence
  and Interactive Digital Entertainment}, Vol.~4. ~1: pp. 216--217. 2008.
\newblock \doi{10.1609/aiide.v4i1.18700}.

\bibitem{pozero}
Grill, Jean-Bastien, Altch{\'e}, Florent, Tang, Yunhao, Hubert, Thomas, Valko,
  Michal, Antonoglou, Ioannis and Munos, R{\'e}mi.
\newblock \enquote{Monte-Carlo tree search as regularized policy optimization.}
\newblock \textit{International Conference on Machine Learning}: pp.
  3769--3778. 2020. PMLR.
\newblock \urlprefix\url{https://proceedings.mlr.press/v119/grill20a.html}.

\bibitem{adam}
Kingma, Diederik~P and Ba, Jimmy.
\newblock \enquote{Adam: A method for stochastic optimization.}
\newblock \textit{arXiv preprint arXiv:1412.6980}
  \doi{10.48550/arXiv.1412.6980}.

\bibitem{rnn}
Rumelhart, David~E, Hinton, Geoffrey~E and Williams, Ronald~J.
\newblock \enquote{Learning internal representations by error propagation.}
\newblock Technical report no.
\newblock California Univ San Diego La Jolla Inst for Cognitive Science.
\newblock 1985.
\newblock \urlprefix\url{https://ieeexplore.ieee.org/document/6302929}.

\bibitem{dwa_shipping2}
Chen, Zheng, Zhang, Youming, Zhang, Yougong, Nie, Yong, Tang, Jianzhong and
  Zhu, Shiqiang.
\newblock \enquote{A Hybrid Path Planning Algorithm for Unmanned Surface
  Vehicles in Complex Environment With Dynamic Obstacles.}
\newblock \textit{IEEE Access} Vol.~7 (2019): pp. 126439--126449.
\newblock \doi{10.1109/ACCESS.2019.2936689}.

\end{thebibliography}

\end{document}